\documentclass[a4paper,twoside]{article}

\usepackage{epsfig}
\usepackage{booktabs}
\usepackage{subcaption}
\usepackage{calc}
\usepackage{amssymb}
\usepackage{amstext}
\usepackage{amsmath}
\usepackage{amsthm}
\usepackage{multicol}
\usepackage{pslatex}
\usepackage{apalike}
\usepackage{algorithm2e}
\usepackage[bottom]{footmisc}
\usepackage{SCITEPRESS}     

\begin{document}

\title{Interpretable Low-Dimensional Modeling of Spatiotemporal Agent States for Decision Making in Football Tactics}

\author{\authorname{Kenjiro Ide\sup{1}, Taiga Someya\sup{2}, Kohei Kawaguchi\sup{3}\orcidAuthor{0000-0003-1058-0048}, Keisuke Fujii\sup{1}\orcidAuthor{0000-0001-5487-4297}}
\affiliation{\sup{1}Graduate School of Informatics, Nagoya University, Japan}
\affiliation{\sup{2}The University of Tokyo, Japan}
\affiliation{\sup{3}The Hong Kong University of Science and Technology, China}
\email{ide.kenjiro@g.sp.m.is.nagoya-u.ac.jp, fujii@i.nagoya-u.ac.jp}
}


\abstract{Understanding football tactics is crucial for managers and analysts. Previous research has proposed models based on spatial and kinematic equations, but these are computationally expensive. Also, Reinforcement learning approaches use player positions and velocities but lack interpretability and require large datasets. Rule-based models align with expert knowledge but have not fully considered all players' states. This study explores whether low-dimensional, rule-based models using spatiotemporal data can effectively capture football tactics. Our approach defines interpretable state variables for both the ball-holder and potential pass receivers, based on criteria that explore options like passing. Through discussions with a manager, we identified key variables representing the game state. We then used StatsBomb event data and SkillCorner tracking data from the 2023$/$24 LaLiga season to train an XGBoost model to predict pass success. The analysis revealed that the distance between the player and the ball, as well as the player’s space score, were key factors in determining successful passes. Our interpretable low-dimensional modeling facilitates tactical analysis through the use of intuitive variables and provides practical value as a tool to support decision-making in football.}

\onecolumn \maketitle \normalsize \setcounter{footnote}{0} \vfill

\section{\uppercase{Introduction}}
\label{sec:introduction}

Understanding football tactics is crucial for managers (head coaches), coaches, and analysts. Previous studies have proposed mathematical models based on spatial evaluation and minimum distance to the ball using geometrical and kinematic equations  \cite{taki1996development,fujimura2005geometric}, but these models are often computationally expensive. Recent reinforcement learning approaches \cite{rahimian2022beyond,nakahara2023action} utilize the positions and velocities of all players but suffer from low interpretability and require large datasets. Rule-based evaluation models address both limited data and alignment with expert knowledge, yet prior work did not consider the states of all players when defining their actions. Furthermore, it is considered that the situation of limited available datasets will continue due to the sports dataset being a licensed business \cite{Lucas2020AllBA}.

This study investigates whether low-dimensional, rule-based state models using spatiotemporal data can effectively capture and understand football tactics. Our approach involves viewing soccer as a game where 11 players from each team compete for time and space. It includes defining interpretable state variables for the ball-holder and multiple potential pass receivers. These state definitions are based on criteria that explore options such as passing. 
Regarding passing actions, researchers have considered modeling and valuing of passes in data-driven manners \cite{Power17,goes2019not,Bransen19,rahimian2022let,rahimian2023pass,robberechts2023xpass}. Unlike these studies, our research adopts an approach that aims to understand decision-making in football tactics by performing interpretable low-dimensional modeling of spatiotemporal agent states in a rule-based manner. We then verify the validity of the modeling and selected features in a data-driven manner.

In this study, we first conducted interviews with a football team manager (head coach) to identify important variables that describe the state of the game. Then, we train a XGBoost model \cite{chen2016xgboost} to predict whether a pass is successful or unsuccessful. The model's performance is validated using the F1 Score on test data. In addition, we qualitatively investigate variables that are important for pass success.

The contributions of this paper are as follows.
First, we define interpretable state variables based on discussions with a football team manager (head coach). This makes it easier for managers and analysts to interpret the results and to make practical use of the results in the analysis of other tactics and players. Second, we found that the distance between the ball and the receiver of the pass was a particularly important factor in the success of the pass compared to other variables. In addition, variables such as the space score of each player and the arrival time of the defensive player closest to the course of the pass were also suggested to influence the success of the pass.

\section{\uppercase{Methods}}

\subsection{Datasets}
We used event data provided by StatsBomb and tracking data provided by SkillCorner for the $2023$/$2024$ season of La Liga. The event data includes the type of action (such as pass or shot), the player who performed it, and when it was performed. Regarding field coordinates, the line connecting the goals was used as the x-axis and, the center line was used as the y-axis. The tracking data stores the position coordinates of all players at all times. We used $200$ of the season's $380$ games. The numbers of successful and unsuccessful passes are $24,416$ and $6,325$, respectively.

\subsection{Preprocessing}
The event data and tracking data do not match the start frames. Therefore, we used the following approach. After fifty frames around the kickoff frame of the event data were obtained, and within that range, the kickoff was defined as four frames before the frame in which the acceleration of the ball was maximum. These four frames were visually confirmed using the tracking data.

For ease of analysis, the direction of attack was aligned to the right and divided by attack sequence. An attack sequence is considered to be identical from the start of a set play or when the ball is won to when the ball is lost. It was assumed that a single failed pass, for example, does not break off the attack, but that the attack continues until there is an effective sequence of opponents. Some attack sequences were eliminated due to a lack of tracking data.

\subsection{Computing variables}
Soccer is a sport where teams compete for ``time'' and ``space''. The attacking team aims to find open spaces before the opposing players can return to defense, quickly developing their attack to get closer to the goal. On the other hand, the defensive team prevents attacks by not giving the opponent time and filling spaces, thereby reducing options for passes and dribbles. We view soccer as a sport where players compete for ``time'' and ``space'' as described above and determine the state variables to express this principle based on feedback from a football manager (or head coach) in Japan.

The state variables have the following structure.
First, it is divided into the relative state and the absolute state.
The relative state is divided into the on-ball state and the off-ball state.
The on-ball state variable is divided into the case that an on-ball action has been taken and the case that no one is taking on-ball action.

In defining these variables, we defined a space score for each player as shown in Figure 1 and the attached video. Each player's space is a Voronoi region that takes speed into account, and we defined the space weight according to the importance of the soccer field: the larger the x-axis, the closer to the opponent's goal, and the closer to $0$ on the y-axis, the closer to the center of the field, the higher the weight. The space score is then calculated by multiplying the area of the Voronoi region by the space weight. The space score can be visualized as shown in Figure 1 and the attached video, with the value of space each player has represented by color. It can be calculated not only for attacking teams but also for defending teams as well. Therefore, it is also possible to represent who the defensive players are that are dangerous if the attacking team loses the ball.

The state variables when the on-ball state action is performed are the distance and angle to the opponent goal, the time taken for the nearest defender to reach it, and the amount of change in the space score when moving $1$ m in $8$ directions (forward, right front, right, right back, back, left back, left, left front). The variables for the on-ball state when no one has the ball are the distance and angle to the opponent's goal of the player closest to the ball on both teams, and the ball speed. The variables for the off-ball state are the space score of each player, the amount of change in space score when each player moves $1$ m in each of the $8$ directions, the distance between the ball and each player, the time each attacking player is closest to the time for the closest defensive player to reach each attacking player, and the time for the closest defensive player to reach the pass line to each attacking player.

In this study, the off-ball state variable was used to determine the score of the passing options for the on-ball player.

\begin{figure}[!h]
  \centering
   {\epsfig{file = 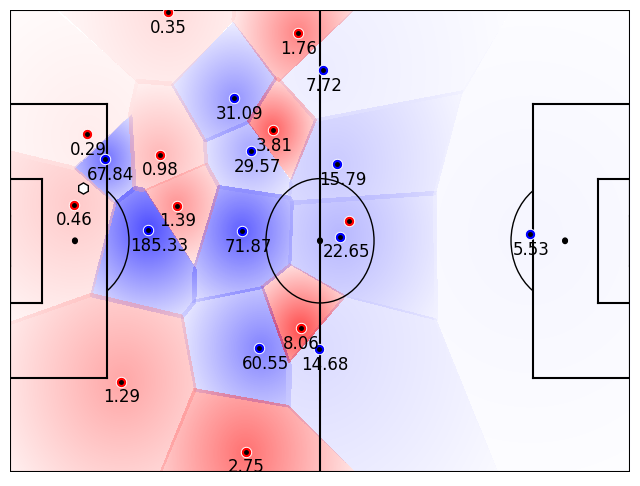, width = 7.5cm}}
  \caption{A visualization of the space score. The red color is the attacking team and the blue color is the defending team. The space score is calculated by multiplying the area of the Voronoi region by the space weight. That of the defensive team is the score multiplied by the area of the Voronoi region and the field weight, which is inverted from left to right as in the case of the attacking team. And the higher the space score, the more likely the team is to be in a position to score. The attached video includes a full version of this attack and the interpretation is described in the Results section. }
  \label{fig:example1}
\end{figure}

\subsection{Modeling}
In this study, the XGBoost model \cite{chen2016xgboost} was used to predict whether an off-ball player's condition would result in a successful or unsuccessful pass. The model was used because it is able to successfully capture non-linear relationships, identify which variables are most influential in predicting successful passes, and analyze factors contributing to the score quantitatively.

The model was built based on variables such as the score of the space in which the off-ball player can move, the distance between the ball and the off-ball player, and the time it takes the closest opponent to reach the off-ball player and the pass line. These four variables are used to calculate the score of the receiver of the pass. This score will also be used in the future to analyze the players' action decisions. For locations with infinite values, interpolation by the median was performed.

In order to focus on the players most likely to influence the goal when predicting the probability of a successful pass using the XGBoost model, we selected only the top $n$ players in each event with the shortest distance with respect to the distance to the ball and the top $n$ players with the highest values for the other variables, and we compared accuracy for each of the four variables selected. The direction with the largest change in space score when moving $1$ m in $8$ directions was used. This approach allowed us to eliminate the influence of players who were unlikely to contribute to the score and to build a model based on more relevant data.

\subsection{Validation}
The model performance was evaluated based on common classification model metrics such as accuracy, repeatability, and F1 score. These metrics were used to quantitatively measure how accurately the model predicts the presence or absence of goals and how it captures the effects of misclassification. In addition, by using $k$-fold cross-validation, the entire dataset was divided into multiple subsets, and the generalization performance of the model was confirmed by repeated training and testing. This prevented over-training and enabled stable performance evaluation.

In the validation, we determined which of the four variables related to calculating the pass receiver's score (the space in which the off-ball player can move, the distance between the ball and the off-ball player, and the time it takes for the nearest opponent to reach either the off-ball player or the passing lane) to use for selecting the top $n$ players by comparing accuracy. Additionally, the value of $n$ for the number of top players selected was set to either $1$ or $3$, based on the coach's opinion that three players are enough for a pass. While increasing $n$ could improve prediction accuracy, we prioritized accurately capturing the decision-making of players contributing to goals. Furthermore, SHAP (Shapley Additive exPlanations) \cite{lundberg2017unified} was employed to identify the features influencing pass success.

\section{Results}

\subsection{Validation of the models}
First, we compared which of the four variables for off-ball players (fast\_space\_vel, the score of space to move; dist\_ball, the distance between the ball and the off-ball player; time\_to\_player, the time it takes for the nearest opponent to reach the off-ball player; time\_to\_passline, the time it takes for the nearest opponent to reach the pass line) would achieve the highest accuracy for the top nth choice. Based on the director's opinion, we fixed $n$ to $3$ and calculated accuracy for fast\_space\_vel to be $0.512$, dist\_ball to be $0.559$, time\_to\_player to be $0.538$, and time\_to\_passline to be $0.521$. Based on these results, dist\_ball, which had the highest accuracy, is used.

Finally, the optimal hyperparameters were determined using Grid Search \cite{liashchynskyi2019grid} with values of $n$ of $1$ and $3$. The results showed that accuracy was $0.685$ for $n$ of $1$ and $0.752$ for $n$ of $3$. Accuracy, precision, recall, and F1 scores were then calculated. The results are shown in Table 1.


\begin{table}[h]
\caption{Evaluation Results for $n=1$ and $n=3$}\label{tab:example1} \centering
\begin{tabular}{ccccc}
  & Accuracy & Precision & Recall & F1 Score \\
  \toprule
  $n=1$ & 0.685 & 0.54 & 0.55 & 0.54 \\
  \midrule
  $n=3$ & 0.752 &  0.58 & 0.55 & 0.56 \\
  \bottomrule
\end{tabular}
\end{table}








\subsection{Qualitative evaluation}
\subsubsection{For important features}
The SHAP values for this model with $n=1$ are shown in Figure 2 and those for $n=3$ in Figure 3.In both cases, the relatively narrow range of SHAP values suggests that all of the top features have a certain impact on the model. In other words, it suggests that the model uses many features in a balanced manner. However, the fact that several of the top SHAP values for dist\_ball are included in both cases indicates that the model places importance on the distance between the off-ball player and the ball. This may capture the tendency for more successful passes to be made at shorter distances. The small SHAP value of the variation\_space\_vel indicates that the amount of change in the space score when the player moves has little effect on the success of the pass.

\begin{figure}[!h]
  \centering
   {\epsfig{file = 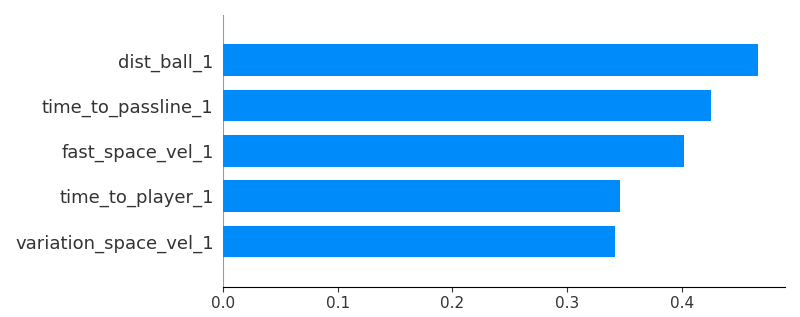, width = 7.5cm}}
  \caption{SHAP for $n=1$. fast\_space\_vel: the score of space to move. dist\_ball: the distance between the ball and the off-ball player. time\_to\_player: the time it takes for the nearest opponent to reach the off-ball player. time\_to\_passline: the time it takes for the nearest opponent to reach the pass line. Variables of the off-ball player with the closest distance to the ball are shown.}
  \label{fig:shap1}
\end{figure}

\begin{figure}[!h]
  \centering
   {\epsfig{file = 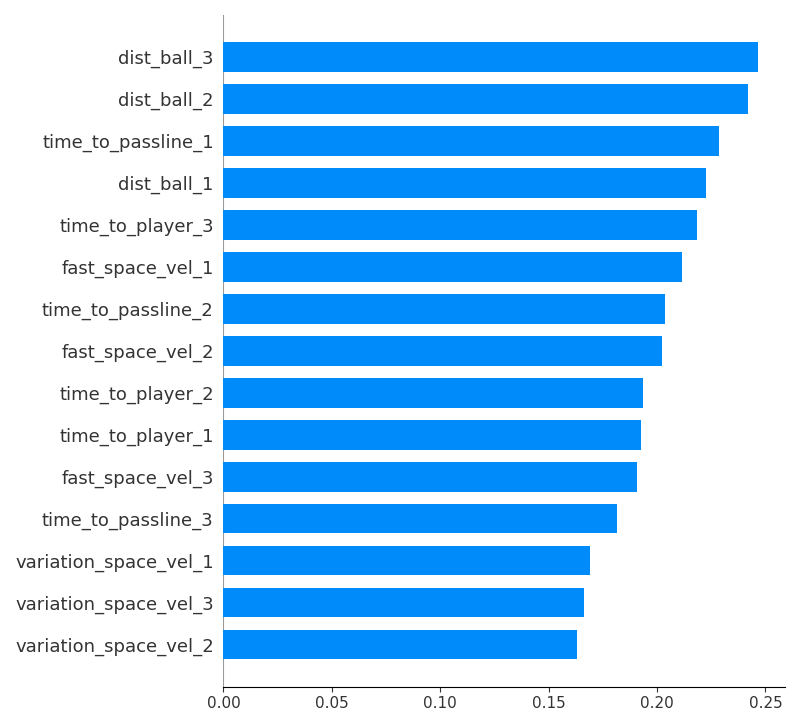, width = 7.5cm}}
  \caption{SHAP for $n=3$. The types of variables are the same as in Figure 2. The variables for the 3 players closest to the ball are shown.}
  \label{fig:shap2}
\end{figure}

\subsubsection{An example video analysis}
We attached an example video visualizing the space score. This is the FC Barcelona vs Real Madrid match played on October $28$, $2023$. The red color represents FC Barcelona and the blue color represents Real Madrid. In this video, FC Barcelona is attacking to the right. If players are offside, they (e.g., rightmost red offensive player) are not included in the calculation. The numbers at the bottom of the players are the space score of each player. Goalkeepers, for example, have more space, but their space score is smaller because they are farther away from the opponent's goal. In this scene, around the 10-second mark, the attacking player with the higher space score moves to get behind the opponent. Then, it can be seen that the players are creating chances by making passes toward the player with the larger space score.

\subsubsection{Limitations of the model}
The XGBoost model used in this study is suitable for evaluating the impact of individual features, but it may struggle to capture interactions between players and tactical coordination. In soccer, the space created by players and their coordination with teammates are crucial factors in passing success, but such complex movements may not be fully captured by XGBoost. Additionally, factors such as the match period (e.g., first or second half) and the game state (e.g., whether the team is winning or losing) can significantly influence player behavior. These contextual influences were not adequately considered in the XGBoost model employed in this study, highlighting a limitation of the approach.









 

\section{\uppercase{Conclusions}}
\label{sec:conclusion}
In this study, we predicted passing success in soccer using variables we defined based on the opinions of the manager. The results showed that the distance between the player and the ball had a relatively large impact on the success of a pass.

The interpretable low-dimensional modeling used in this study made it easier to interpret the results by using intuitively understandable variables such as the distance between players and the ball, and space scores, in predicting pass success. Furthermore, the high interpretability allows for feedback from managers and analysts, potentially contributing to tactical improvements. Therefore, this interpretable low-dimensional model is considered to have practical value as a support tool for understanding and decision-making in tactical analysis on the field.

Although only the four off-ball player variables were used, future studies will extend the analysis by adding the on-ball player state variables and the absolute state described in section 2.3. Furthermore, by using reinforcement learning \cite{nakahara2023action} and adding other state and action modeling, we can analyze not only passing situations, but also a wider range of situations, including shooting phases, off-ball player actions, and defensive analysis.

\bibliographystyle{apalike}
{\small
\bibliography{example}}

\end{document}